\documentclass[letterpaper, 10 pt, conference]{ieeeconf}  

\IEEEoverridecommandlockouts                              

\usepackage{graphics} 
\usepackage{epsfig} 
\usepackage{mathptmx} 
\usepackage{times} 
\usepackage{amsmath} 
\usepackage{amssymb}  
\usepackage{booktabs}
\usepackage{multirow}
\usepackage{svg}
\newcommand{\textbfr}[1]{\textcolor{red}{\textbf{#1}}}
\newcommand{\textbl}[1]{\textcolor{blue}{#1}}

\usepackage{color}

\usepackage{fancyhdr}

\title{\LARGE \bf
CAPT: Category-level Articulation Estimation from a Single Point Cloud Using Transformer
}

\author{Lian Fu$^{1}$, Ryoichi Ishikawa$^{1}$, Yoshihiro Sato$^{2}$ and Takeshi Oishi$^{1}$
\thanks{$^{1}$The authors are with The Institute of Industrial Science, The University of Tokyo, Japan.
        {\tt\small \{lianfu, ishikawa, oishi\}@cvl.iis.u-tokyo.ac.jp}}%
\thanks{$^{2}$The author is with Faculty of Engineering, Kyoto University of Advanced Science, Japan.
        {\tt\small sato.yoshihiro@kuas.ac.jp}}%
}

\begin{document}

\maketitle
\thispagestyle{fancy}
\fancyhead{} 
\fancyfoot{} 
\fancyfoot[C]{
\small
\parbox{54em}{
© 2024 IEEE.  Personal use of this material is permitted.  Permission from IEEE must be obtained for all other uses, in any current or future media, including reprinting/republishing this material for advertising or promotional purposes, creating new collective works, for resale or redistribution to servers or lists, or reuse of any copyrighted component of this work in other works.}
}


\begin{abstract}
The ability to estimate joint parameters is essential for various applications in robotics and computer vision. 
In this paper, we propose CAPT: category-level articulation estimation from a point cloud using Transformer. 
CAPT uses an end-to-end transformer-based architecture for joint parameter and state estimation of articulated objects from a single point cloud.
The proposed CAPT methods accurately estimate joint parameters and states for various articulated objects with high precision and robustness. 
The paper also introduces a motion loss approach, which improves articulation estimation performance by emphasizing the dynamic features of articulated objects. 
Additionally, the paper presents a double voting strategy to provide the framework with coarse-to-fine parameter estimation. 
Experimental results on several category datasets demonstrate that our methods outperform existing alternatives for articulation estimation. 
Our research provides a promising solution for applying Transformer-based architectures in articulated object analysis.

\end{abstract}
\section{Introduction}

\begin{figure}[tb]
	\begin{center}
		\includegraphics[width=8.0cm]{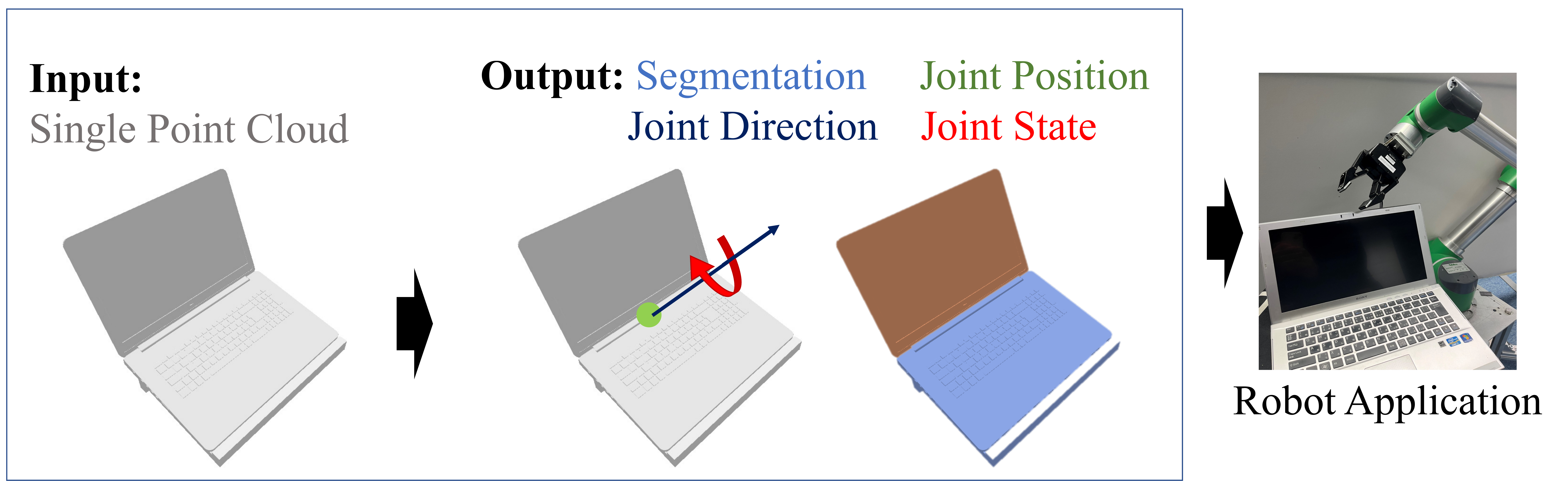}
			\caption[labelInTOC]{Articulation estimation aims to estimate joint parameters and states from visual information. In our case, we propose to infer from only a single static point cloud. This task could be applied in virtual/augmented reality, robot interaction, etc.}
		\label{intro}
	\end{center}
\end{figure}

Accurate articulation estimation is essential for a wide range of applications in robotics. Articulation estimation is the task of obtaining the joint parameters and joint states of the objects from visual input, as depicted in Fig. \ref{intro}. With the development of Deep Learning, related methods have now shifted from instance level toward estimating category-level articulation parameters from videos \cite{jainScrewNetCategoryIndependentArticulation2021} or pairs of images \cite{jiangDittoBuildingDigital2022}. 

Even so, category-level articulation estimation based on a single static point cloud remains a challenging task.
The category prior provides only an ambiguous indication of an object's kinematic constraints, instead of clear information about the spatial structure as in a CAD model. Moreover, as dynamic properties, articulation parameters are not explicitly stored in static point clouds. 
Previous research has commonly approached this task by dividing it into multiple stages \cite{wangShape2MotionJointAnalysis2019a} or incorporating post-optimization algorithms \cite{liCategorylevelArticulatedObject2020}. However, such methods require relatively complex training procedures. The performance of a latter-stage model relies on a corresponding former-stage model, where a poor result in the former stage usually causes an even worse final estimation \cite{jiangDittoBuildingDigital2022}.

To address these challenges, we propose CAPT: category-level articulation estimation based on a single point cloud using Transformer. 
We introduce an end-to-end Transformer-based architecture for articulation estimation based on a single point cloud. 
To emphasize the dynamic nature of articulated objects, we also propose a motion loss approach to restore the dynamic features. 
Additionally, we design a high-accuracy double voting strategy to determine the final predicted parameter values. 
To the best of our knowledge, this is the first work that utilizes a Transformer-based architecture for the estimation of joint parameters of articulated objects.

Experimental results from several category datasets demonstrate that our methods achieve better performance than previously published alternatives for articulation estimation. 
Our methods accurately estimate the joint parameters of various articulated objects with high precision and robustness. 
This work opens new opportunities for the application of Transformer-based architectures in the field of articulated object analysis and control.

The main contributions of this paper are summarized as follows: 
\begin{itemize}
\item Proposal of an end-to-end category-level articulation estimation model using Transformer
\item Proposal of a motion loss approach that improves articulation estimation performance by emphasizing the dynamic character of articulated objects
\item Design of a high-accuracy double voting strategy to decide the final predicted parameters 
\item Experiments on a synthetic dataset to demonstrate the high accuracy of our methods
\end{itemize}

\section{Related Work}
\subsection{Articulation Estimation}
Existing articulation estimation methods can be grouped into three categories based on the input type: interaction-based, multi-view-based, and single-view-based. Interaction methods provide sufficient dynamic information for articulation estimation, as made clear in numerous studies \cite{hausmanActiveArticulationModel2015}\cite{sturmProbabilisticFrameworkLearning2011}\cite{bohgInteractivePerceptionLeveraging2017}\cite{katzInteractiveSegmentationTracking2013}. Related methods focus on using the differences in the observations before and after the interaction to help estimate an initial guess of the articulation model \cite{hartantoHandMotionguidedArticulationSegmentation2020} or build up an implicit neural-representation of the object \cite{jiangDittoBuildingDigital2022}. Multi-view methods infer the articulations from multiple observations \cite{yiDeepPartInduction2018}\cite{colleoniDeepLearningBased2019}\cite{heppertCategoryIndependentArticulatedObject2022a}, with recent works including ScrewNet \cite{jainScrewNetCategoryIndependentArticulation2021}, CAPTRA \cite{wengCAPTRACAtegorylevelPose2021a}, DUST-net \cite{jainDistributionalDepthBasedEstimation2022} and CLA-NeRF \cite{tsengCLANeRFCategoryLevelArticulated2022}. Compared with the other two types of methods, single-view methods require as little information as a single point cloud or a depth image \cite{yanRPMNetRecurrentPrediction2019}\cite{liuRealWorldCategorylevelArticulation2022}\cite{liuSemiWeaklySupervisedObject2023}. To achieve single-view estimation, existing methods usually approach this task with multi-stage networks. RPM-Net \cite{yanRPMNetRecurrentPrediction2019} first predicts a temporal sequence of pointwise displacements using recurrent neural network before estimating articulation. ANCSH \cite{liCategorylevelArticulatedObject2020} predicts the articulation parameters with the help of articulation-aware normalized coordinate space hierarchy and post-optimization. Related research also combines manipulation with articulation estimation, including FlowBot3D \cite{eisnerFlowBot3DLearning3D2022}. Recently, Liu et al. \cite{liuSemiWeaklySupervisedObject2023} proposed a semi-weakly supervised approach using a Graph Neural Network (GNN) which takes pre-segmented point clouds as input. While single-view methods seem to be the most useful and practical, their performance remains in need of improvement.

\subsection{Transformer for Point Clouds}
Ever since its proposal by Vaswani et al. \cite{vaswaniAttentionAllYou2017}, Transformer has become the most popular neural model not only in natural language processing (NLP) but also in computer vision (CV). 
Vision-Transformer (ViT) \cite{dosovitskiyImageWorth16x162020} brought the Transformer into vision tasks and pushed performance in many CV tasks to new heights. The permutation-equivariance of the self-attention mechanism fits the inherently unordered point cloud data structure well. 
Transformer allows for the capture of global contextual information, compared with the limited receptive field in convolutional methods. Such global capture can be useful in understanding the overall structure of the point cloud. Guo et al. \cite{guoPctPointCloud2021} and Zhao et al. \cite{zhaoPointTransformer2021} proposed Point Cloud Transformer and Point Transformer, respectively; both demonstrate the potential of Transformer in the field of Deep Vision. Transformer has been adopted in various point cloud processing tasks, including point cloud completion \cite{linPCTMANetPointCloud2021}, denoising \cite{xuTDNetTransformerbasedNetwork2022}, and registration \cite{wangDeepClosestPoint2019}\cite{yuPointBERTPretraining3D2022}. We believe that it is appropriate to apply an attention-based approach to articulation estimation. 

\begin{figure*}[t]
	\begin{center}
 \vspace{5mm}
		\includegraphics[width=1\linewidth]{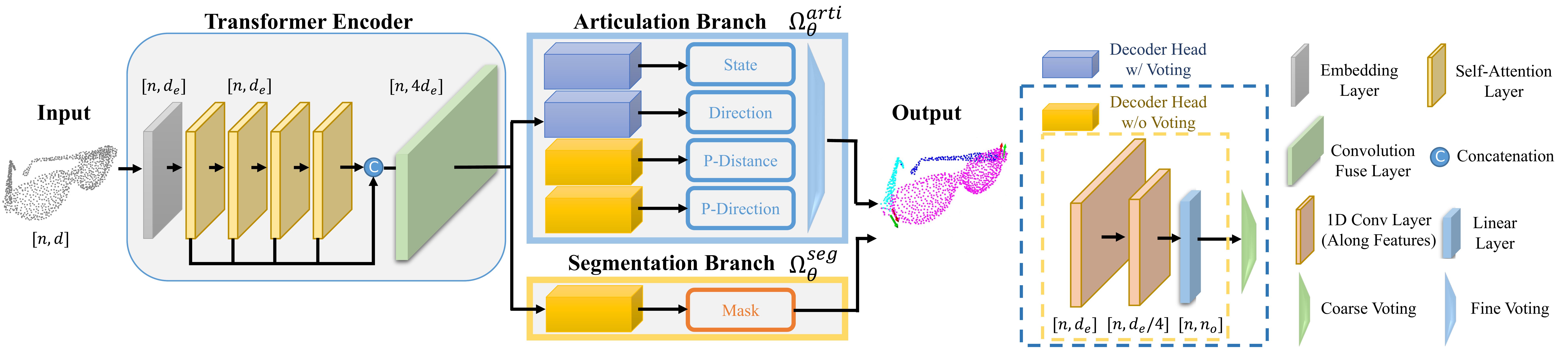}
			\caption[labelInTOC]{Our CAPT (category-level articulation estimation from a point cloud using Transformer) architecture. $n$ is the number of points, $d$ is the origin feature dimension of each point and $d_\mathrm{e}$ is the embedded feature dimension. In the output, as in all figures in this paper, the red arrow represents the predicted joint, while the green arrow represents the ground truth joint.}
		\label{model}
	\end{center}
\end{figure*}

\section{Category-level Articulation Estimation Framework}
In this section, we first clarify the problem formulation. 
Next, we introduce the structure of our framework, including a Transformer-based encoder and a multi-branch decoder. 
The encoder is identical to that of PCT \cite{guoPctPointCloud2021}, but for better understanding, we briefly explain it in Sec. \ref{ssec_embedding}.
The overall architecture of CAPT is shown in Fig. \ref{model}.

\subsection{Problem Formulation}
Given an articulated object $O$ from a known category with $n_{\mathrm{J}}$ joints and $n_{\mathrm{L}}$ links, a partial point cloud with $n$ points $Q = \{\mathbf{p}_i\in \mathbb{R}^{d} | i=1,\dots,n\}$ may be obtained from a single observation. 
The dimension $d=3$ if the space coordinates $x, y, z$ are used for each point, which is the case in our method. However, $d$ could certainly be higher if more information is added to each point, such as color space $r, g, b$ (where $d=6$) \cite{parkColoredPointCloud2017} or normal vector space $v^x, v^y, v^z$ (where $d=6$) \cite{wuPointConvDeepConvolutional2019}. 

The expected outputs of the method are the segmentation of $Q$ into $n_{\mathrm{L}}$ links, parameters, and states for all $n_{\mathrm{J}}$ joints. 
Point-wise segmentation predicts a segmentation label for each point $C = \{c(\mathbf{p}_i)\in {1,\dots,n_{\mathrm{L}}} | i=1,\dots,n\}$. 
Here, joints with one degree of freedom (DoF) are considered, 
since they are the most common joints. 
For joint $J_k
(k=1,...,n_{\mathrm{J}})$, one property is joint direction, denoted by unit vector ${\phi}_k^{\mathrm{dir}}\in \mathbb{R}^3$. 
Although joint axes are usually considered to be directionless, we manually define a direction for each joint under a consistent rule. 
We also consider joint position, denoted as a point on the axis named pivot ${\phi}_k^{\mathrm{pivot}}\in \mathbb{R}^3$. 
Joint state 
is represented in radians, ${\phi}_k^{\mathrm{s}}\in (-\pi, \pi]$, and assigned a predefined zero state for each category. 

\subsection{Input embedding and Encoder}
\label{ssec_embedding}
As a disordered data structure, a point cloud needs no position embedding as does Transformer in NLP \cite{vaswaniAttentionAllYou2017} or patch embedding as does ViT \cite{dosovitskiyImageWorth16x162020}. However, as discussed in \cite{qiPointnetDeepLearning2017}, adopting local feature embedding and neighbor feature embedding improves feature extraction performance. The encoder aims to extract high-dimensional features from 3D point clouds, so as to provide sufficient information for attached decoders to estimate targets. Our model adopts the same encoder as \cite{guoPctPointCloud2021}. The encoder consists of four self-attention layers, which are used to process the point cloud data in a hierarchical manner. A feature map of $F^\mathrm{e}\in \mathbb{R}^{n\times 4d_\mathrm{e}}$ is finally outputted by the encoder and fed into the decoder.

\subsection{Multi-branch Decoders}
Multi-branch decoders, attached downstream from the encoder, execute several tasks. 
The decoders consist of articulation branch $\Omega_\mathrm{\theta}^\mathrm{arti}$ and segmentation branch $\Omega_\mathrm{\theta}^\mathrm{seg}$. 
The output channel number is determined by the category property, including the part number, joint number, and joint type for each sample in the category dataset. 
Within a given category, the link number and joint number are not always fixed. 

Here we adopt a fully prepared strategy, which considers the number of parts $n_{\mathrm{L}}$ and the number of joints $n_{\mathrm{J}}$ for the output channels to represent the maximum number for the whole category dataset. 
For those samples whose numbers of parts or joints are below the maximum, we set the output of the extra channels to zero.

The segmentation branch predicts a point-wise segmentation mask for each link, i.e.,
\begin{equation}
\Omega^\mathrm{seg}_\mathrm{\theta}(F_\mathrm{e})\rightarrow \hat{H}(Q)\in \mathbb{R}^{n\times n_\mathrm{L}},
\end{equation}
where $\hat{H}(Q)$ is the possibility distribution for each point in $Q$ belonging to each part.
%
%
%
The joint axis direction may be defined by a unit direction vector $\phi_k^\mathrm{dir} \in \mathbb{R}^3$. 
The joint position may be represented by any point in the point cloud $\textbf{p}_i\in Q$; a unit vector $\phi_k^\mathrm{pdir}(\mathbf{p}_i)\in \mathbb{R}^{3}$ starting from $\mathbf{p}_i$ and perpendicular to the joint $J^{\mathrm{r}}_k$; and the distance $\phi_k^\mathrm{pdir}(\mathbf{p}_i)\in \mathbb{R}^{3}$ between $\textbf{p}_i$ and $J^{\mathrm{r}}_k$.

To take better advantage of the global receptive field of Transformer, the articulation branch conducts a point-wise prediction of all articulation parameters:
%
\begin{eqnarray}
&\Omega_\mathrm{\theta}^\mathrm{arti}
(F_\mathrm{e})\rightarrow \hat{\Phi}_k(\textbf{p}),\\
&\hat{\Phi}_k(\textbf{p})=\{\hat{\mathrm{\phi}}_k^{\mathrm{dir}}(\textbf{p}), \hat{\phi}_k^{\mathrm{dist}}(\textbf{p}), \hat{\phi}_k^{\mathrm{pdir}}(\textbf{p}), \hat{\phi}_k^{\mathrm{s}}(\textbf{p})\}.
\end{eqnarray}
%
$\hat{\Phi}_k(\textbf{p})$ consists of joint direction $\hat{\phi}_k^{\mathrm{dir}}(\textbf{p})\in \mathbb{R}^{3}$, point-to-joint distance $\hat{\phi}_k^{\mathrm{dist}}(\textbf{p})\in \mathbb{R}$, point-to-joint direction $\hat{\phi}_k^{\mathrm{pdir}}(\textbf{p})\in \mathbb{R}^{3}$, and joint state $\hat{\phi}_k^{\mathrm{s}}(\textbf{p})\in \mathbb{R}^{3}$. 

The predicted joint pivot of each point is computed as
\begin{equation}
\hat{\phi}_k^{\mathrm{pivot}}(\textbf{p})=\textbf{p}+\hat{\phi}_k^{\mathrm{dist}}(\textbf{p})\cdot \hat{\phi}_k^{\mathrm{pdir}}(\textbf{p}).
\end{equation}

\begin{figure}[tb]
	\begin{center}
		\includegraphics[width=0.9\linewidth]{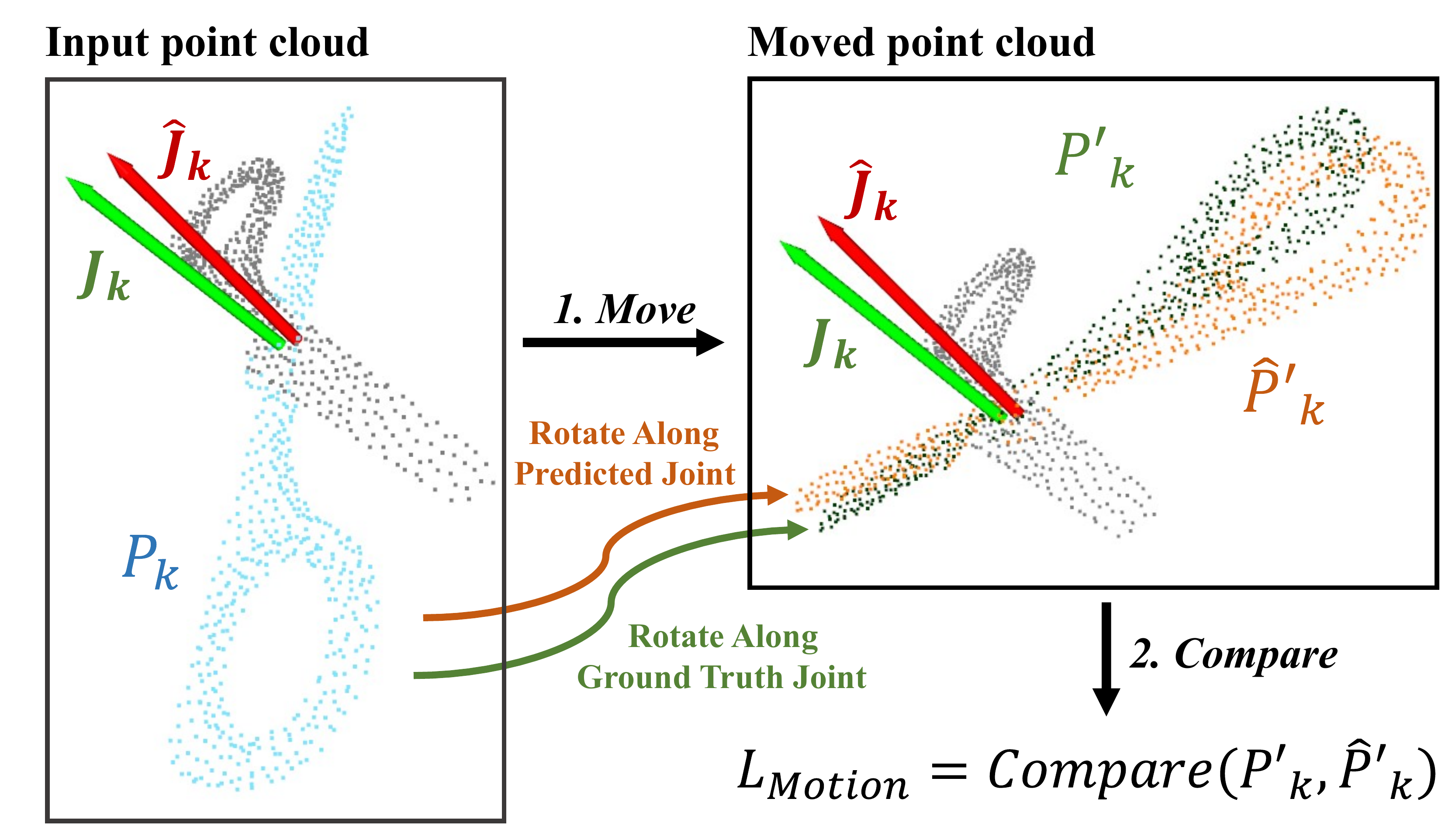}
			\caption[labelInTOC]{Diagram of motion loss calculation. Motion loss of $k^{th}$ joint is calculated in two steps. (1) Move: Moving the part point cloud $P_k$ along predicted joint $\hat{J}_k$ and ground truth joint $J_k$ to obtain rotated point clouds $\hat{P}'_k$ and $P'_k$, respectively. (2) Compare: Get motion loss by comparing $\hat{P}'_k$ and $P'_k$. The total motion loss is the sum of each joint's motion loss.}
		\label{motion}
	\end{center}
\end{figure}

\section{Loss Design and Optimization}
We next explain the proposed motion loss approach, which recovers dynamic features of joints from static input. 
We also describe the total loss design and the double voting strategy, which inherits the coarse-to-fine inference paradigm in estimating articulation model parameters. 
\subsection{Motion Loss}
Articulation is a dynamic property of objects that is not explicitly expressed by a single static point cloud. 
However, restoring explicit dynamic information is made possible by using additional constraints. 
The intuitive idea behind motion loss is to move the point cloud of parts separately according to the predicted and ground-truth joints and then compare the moved point clouds.
A diagram of this calculation is shown in Fig. \ref{motion}.

Formally, given point cloud $Q=\{\textbf{p}_i\in \mathbb{R}^3|i=1,\dots ,n\}$, the model provides a prediction of the joint direction $\hat{\phi}_k^{\mathrm{dir}}$ and pivot $\hat{\phi}_k^{\mathrm{pivot}}$ of a joint $J_k$. 
We also know the ground truth direction $\phi_k^{\mathrm{dir}}$, pivot position $\phi_k^{\mathrm{pivot}}$ of $J_k$, and segmentation label $C$ for loss computation. 
We first obtain the part point cloud for the $k_{th}$ link $P_k=\{\mathbf{p}_i \text{ if } C(\mathbf{p}_i)=k\}$, which is the child link of $J_k$. 
Denote the point number of $P_k$ as $n_k$. 
A rotated point cloud $P'_k$ may be obtained by using the Rodrigues rotation formula \cite{chengHistoricalNoteFinite1989}, which rotates points $P_k$ along an arbitrary 3D axis decided by direction $\phi_k^{\mathrm{dir}}$ and any point on the axis $\phi_k^{\mathrm{pivot}}$, by an amount $\alpha$, i.e.,
\begin{equation}
P'_k = 
\Psi_\mathrm{Rodrigues}
(P_k, \phi_k^{\mathrm{dir}}, \phi_k^{\mathrm{pivot}}, \alpha).
\end{equation}
We set rotation angle $\alpha$ as $\pi/2$. $\hat{P}'_k$ may be obtained similarly by rotating the part point cloud $P_k$ along predicted joint axis. The motion loss is then calculated as
\begin{equation}
\label{motion_loss}
    L_{\mathrm{motion}}(P'_k, \hat{P}'_k) = \sum_k^{n_\mathrm{L}}\sum_i^{n_k}\frac{\Vert \mathbf{p}'_{k,i} -\hat{\mathbf{p}}'_{k,i}\Vert}{n_\mathrm{L}\cdot n_k},
\end{equation}
where $\mathbf{p}'_{k,i}\in P'_k, \hat{\mathbf{p}}'_{k,i}\in \hat{P}'_k$. In our case, we choose the least squared error loss (L2 loss) as the per-point space distance function
$Compare$.

\subsection{Total Loss}
The total loss is composed of the losses for segmentation, joint direction, joint position, and joint state. 
For segmentation, we compute the cross-entropy loss as
\begin{equation}
L_{\mathrm{seg}}=-\sum_{i}^{n}\sum_{k}^{n_\mathrm{L}} y_i \cdot \log(\hat{H}(\mathbf{p}_i)),
\end{equation}
where $y_i=1$ if $C(\mathbf{p}_i)=k$, otherwise $y_i=0$.

For joint direction prediction, we compute the cosine similarity distance as
\begin{equation}
L_{\mathrm{dir}} = \sum_k^{n_\mathrm{J}}\sum_i^{n_k}\frac{
1-
\Psi_\mathrm{cossim}(\hat{\phi}_k^{\mathrm{dir}}(\mathbf{p}_i), \phi_k^{\mathrm{dir}}(\mathbf{p}_i))
}{n_\mathrm{J}\cdot n_k},
\end{equation}
where 
$\Psi_\mathrm{cossim}(u, v)$
represents the cosine similarity between $d$-$dim$ vectors $u$ and $v$ as
\begin{equation}
\Psi_\mathrm{cossim}(u ,v) = \frac{\sum_{i=1}^{d}u_i \cdot v_i}{\sqrt{\sum_{i=1}^{d}u_i^2} \cdot \sqrt{\sum_{i=1}^{d}v_i^2}}.
\end{equation}
For joint position prediction, we again compute the cosine similarity distance as
\begin{equation}
L_{\mathrm{pdir}} = \sum_k^{n_\mathrm{J}}\sum_i^{n_k}\frac{1-
\Psi_\mathrm{cossim}
(\hat{\phi}_k^{\mathrm{pdir}}(\mathbf{p}_i), \phi_k^{\mathrm{pdir}}(\mathbf{p}_i))}
{n_\mathrm{J}\cdot n_k},
\end{equation}
and the L2 loss for point-to-joint distance as
\begin{equation}
L_{\mathrm{dist}} = \sum_k^{n_\mathrm{J}}\sum_i^{n_k}\frac{\Vert\hat{\phi}_k^{\mathrm{dist}}(\mathbf{p}_i)- \phi_k^{\mathrm{dist}}(\mathbf{p}_i)\Vert}
{n_\mathrm{J}\cdot n_k},
\end{equation}
Joint state prediction loss is computed as least absolute deviations loss (L1 loss), i.e.,
\begin{equation}
L_{\mathrm{s}} = \sum_k^{n_\mathrm{J}}\sum_i^{n_k}\frac{|\hat{\phi}_k^{\mathrm{s}}(\mathbf{p}_i) - \phi_k^{\mathrm{s}}(\mathbf{p}_i)|}
{{n_\mathrm{J}}\cdot n_k}.
\end{equation}

The overall loss is computed as
\begin{equation}
L=[L_{\mathrm{seg}}, L_{\mathrm{dir}}, L_{\mathrm{pdir}}, L_{\mathrm{dist}}, L_{\mathrm{s}}, L_{\mathrm{motion}}]\cdot \textbf{W}^\mathrm{T}.
\end{equation}
We roughly set the weight combination $\textbf{W}$ used in this paper to be $\textbf{W}=[1,1,1,1,1,0.1]$, considering a balanced learning rate for each target.

\subsection{Double Voting}
The decoder gives per-point prediction $\hat{\Phi}_k(\textbf{p})$ for the parameters and state of joint $J_k$. Points at different distances from $J_k$ usually contain different amounts of information; such differences should not be ignored. Furthermore, points around or on the axis give poor predictions, since the normals of such points may be unstable due to the articulation structure. In order to rule these points out, we propose double voting, a coarse-to-fine voting strategy for accurately and robustly deciding the target estimated parameter values. As in the operation described in \cite{jiangDittoBuildingDigital2022}, a coarse voting is first conducted, i.e.,
\begin{equation}
\label{vote}
\begin{aligned}
    \bar{\Phi}_k
    =\sum_i^{n}\frac{\hat{\Phi}_k(\mathbf{p}_i)}{n},
\end{aligned}
\end{equation}
where all points are weighted equally. Coarse voting assumes that all points contain the same amount of information about $J_k$, but this assumption holds only if the object lies fully along the joint axis. To focus on the points that probably contain more information and are more likely to give good predictions of joint parameter values, we conduct a subsequent round of fine voting. We first compute the distance to the coarsely voted joint $J_k$ for each point as
 \begin{equation}
 \hat{\phi}_k^{\mathrm{dist}}(\textbf{p})=
 \Psi_\mathrm{dist}
 (\textbf{p}, \hat{\phi}_k^{\mathrm{pivot}}, \hat{\phi}_k^{\mathrm{dir}}),
 \end{equation}
 where
\begin{equation}
\Psi_\mathrm{dist}
(\textbf{p},\textbf{q},\textbf{u})=\Vert ((\textbf{p}-\textbf{q})\cdot \textbf{u}^\mathrm{T})\cdot \textbf{u} - (\textbf{p}-\textbf{q})\Vert.
\end{equation}
We set a simple distance threshold, i.e.,
\begin{equation}
\textbf{p}_i\rightarrow Q_k^{\mathrm{f}} \text{ if } \hat{\phi}_k^{\mathrm{dist}}(\mathbf{p}_i) \in [\omega_{0},\omega_{1}] \cdot\beta(\hat{\Phi}_k^{\mathrm{dist}}),
\end{equation}
where $Q_k^{\mathrm{f}}$ is a subset of $Q$ after removing points far from estimated joint $\hat{J}_k$. $\beta(\textbf{x})$ indicates the median value of $\textbf{x}$. $\omega_{0}$ and $\omega_{1}$ are hyperparameters for adjusting the range of points involved in fine voting. A larger $\omega_{0}$ value means that more distant points also contribute to the final result, while a smaller value means that only very close points participate in the fine-voting round. $\omega_{1}$ performs just the opposite role, controlling how close a point may be and still being counted in the fine voting round. Finally, the estimated parameters after fine voting are computed as
\begin{equation}
\bar{\Phi}_k^{\mathrm{f}}
=\sum_{q\in Q_k^{\mathrm{f}}}\frac{\hat{\Phi}_k(\textbf{q})}{n_k^{\mathrm{f}}},
\end{equation}
where $n_k^{\mathrm{f}}$ is the number of points in $Q_k^{\mathrm{f}}$.

\section{Experiments}
\subsection{Datasets}
Several articulated object datasets have been proposed recently \cite{liuRealWorldCategorylevelArticulation2022}\cite{michel2015pose}\cite{xiangSAPIENSimulAtedPartbased2020a}\cite{huLearningPredictPart2017}, but Shape2Motion \cite{wangShape2MotionJointAnalysis2019a} is still the most used and most appropriate for articulation estimation tasks. 
We used four subsets for quantitative evaluation.
We also conducted a qualitative evaluation of the other subsets to demonstrate that our methods could achieve good performance on other synthetic categories. 

Here we briefly summarize the data used in our experiment. After considering the diversity, representativeness, and feasibility of the test set, we selected $\bf{laptop}$, $\bf{washing machine}$, $\bf{oven}$, and $\bf{eyeglasses}$ for quantitative evaluation experiments. $\bf{Scissors}$ and $\bf{bike}$ were also used for qualitative experiments. The joint numbers of these objects (1, 1, 1, 2, 2, and 4, respectively) cover various kinematic structures. The numbers of models in each category are 86, 62, 42, 43, 26, and 63, respectively.
 
Point clouds of articulated objects with annotation labels were generated through simulation in Pybullet \cite{coumans2021}. 
Each joint was set to random initial states within the motion limits. Data were further augmented through random rotation along the $x$-axis, $y$-axis, and $z$-axis, with $R_x, R_y, R_z \in [-\pi, \pi]$; random translation along each axis with range $[-1, 1]$; and scaling by ratios with range $[0.8, 1.2]$. 
We split the dataset into training, validation, and test data sets with a volume ratio of 7:2:1, respectively. 
We ensured that all inputs were unseen objects with random states and points of view during testing.

\subsection{Baselines}
We compared our methods with a simple PCT approach that combines CAPT-plain and ANCSH \cite{liCategorylevelArticulatedObject2020}. 
This simple PCT approach shares the same encoder structure as our method but uses a single-branch multilayer perceptron as the decoder and directly regresses the joint parameters.
This approach was used as a control to determine whether a powerful Transformer encoder alone is sufficient for satisfactory completion of the articulation estimation task.
ANCSH also estimates category-level articulation parameters from single-point-cloud input. However, ANCSH requires a post-optimization process, unlike our end-to-end method.
To determine the effects of motion loss and double voting, we also used CAPT-plain for comparison, which is CAPT without motion loss and double voting.

\subsection{Evaluation Metrics}
We used the following metrics to evaluate our method. We used per-point accuracy (PA) and mean intersection over union (mIoU) to evaluate the segmentation result for each point. For joint position, we use average Euclidean distance (AED) between the predicted joint axis and the ground truth joint axis. For joint direction, we used mean error in degrees (MED) between the predicted joint axis direction and the ground truth joint axis direction. We again used mean error in degrees to evaluate joint state estimation accuracy. To assess the stability of the evaluated method, we used the proportions of results with joint position error values below 0.05 and below 0.01 (reported as AP5 and AP10, respectively, under the joint position columns), the proportions of results with joint direction error below 5 degrees and below 10 degrees (reported as AP5 and AP10, respectively, under the joint direction columns), and the proportions of results with joint state error values below 5 degrees and below 10 degrees (reported as AP5 and AP10, respectively, under the joint state columns).

\begin{table*}[t]
\centering
\setlength{\tabcolsep}{0.6mm}
\vspace{5mm}
\caption[]{Evaluation results for four test data sets: \textup{\textbf{PA} refers to per-point accuracy. \textbf{mIoU} refers to mean intersection over union. \textbf{AP}$x$ refers to average precision (the proportion of results with error below $x$). \textbf{MED} refers to mean error in degrees. \textbf{AED} refers to average Euclidean distance. Values for each joint are represented as multiple values in a single cell for the multiple joints category.}}
\begin{tabular}{@{}cclllllllllll@{}}
\toprule
\multirow{2}{*}{Dataset} & \multirow{2}{*}{Method} & \multicolumn{2}{c}{Segmentation} & \multicolumn{3}{c}{Joint Direction} & \multicolumn{3}{c}{Joint Position}& \multicolumn{3}{c}{Joint State}\\&& \multicolumn{1}{c}{AP$\uparrow$} & \multicolumn{1}{c}{mIoU$\uparrow$} & \multicolumn{1}{c}{MED$\downarrow$} & \multicolumn{1}{c}{AP5$\uparrow$} & \multicolumn{1}{c}{AP10$\uparrow$} & \multicolumn{1}{c}{AED$\downarrow$} & \multicolumn{1}{c}{AP1$\uparrow$} & \multicolumn{1}{c}{AP5$\uparrow$} & \multicolumn{1}{c}{MED$\downarrow$} & \multicolumn{1}{c}{AP5$\uparrow$} & \multicolumn{1}{c}{AP10$\uparrow$} \\
\midrule

\multirow{4}{*}{Eyeglasses}
& PCT & - & - & 7.74, 6.98 & 0.29, 0.38 & 0.73, 0.79 & 0.05, 0.04 & 0.12, 0.14 & 0.53, 0.60 & 11.3, 11.1 & \textbl{0.27}, \textbl{0.26} & \textbl{0.52}, \textbl{0.50}\\
& ANCSH & 0.89 & 0.73   & \textbl{6.35}, 6.37 & \textbl{0.48}, \textbl{0.49} & \textbl{0.84}, 0.84  & 0.09, 0.11 & 0.08, 0.01 & 0.37, 0.27 & 13.9, 13.2 & 0.04, 0.05 & 0.11, 0.12  \\
& CAPT-Plain & \textbfr{0.97}& \textbfr{0.90}  & 6.86, \textbl{5.83} & 0.35, 0.48 & 0.83, \textbl{0.90}  & \textbl{0.04}, \textbl{0.04} & \textbl{0.15}, \textbl{0.17} & \textbl{0.62}, \textbl{0.70} & \textbl{10.14}, \textbl{10.49}  & 0.26, 0.26 & 0.52, 0.49  \\
& CAPT (ours)  & \textbl{0.95} & \textbl{0.87}   & \textbfr{5.15, 4.26}     & \textbfr{0.54, 0.72}     & \textbfr{0.95, 0.96}      & \textbfr{0.03, 0.03}     & \textbfr{0.21, 0.20}     & \textbfr{0.81, 0.82}     & \textbfr{6.58, 6.84}     & \textbfr{0.51, 0.49}     & \textbfr{0.79, 0.79}      \\ \midrule
\multirow{4}{*}{Laptop}
& PCT & -    & -      & 5.40  & 0.54  & 0.94   & 0.02  & 0.30  & 0.92  & 10.90 & 0.26  & 0.51   \\
& ANCSH & 0.52 & 0.33   & 9.06  & 0.48  & 0.71   & 0.50  & 0.01  & 0.01  & 12.10 & 0.02  & 0.03   \\
& CAPT-Plain & \textbl{0.98} & \textbl{0.95}   & \textbfr{2.13} & \textbfr{0.95} & \textbfr{0.99}  & \textbl{0.01}  & \textbl{0.50}  & \textbl{0.98}  & \textbfr{9.38} & \textbl{0.30} & \textbfr{0.55} \\
& CAPT (ours) & \textbfr{0.98}& \textbfr{0.95} & \textbl{2.20}  & \textbl{0.95}  & \textbl{0.99} & \textbfr{0.01} & \textbfr{0.53} & \textbfr{0.98} & \textbl{9.80}  & \textbfr{0.59}  & \textbl{0.54} \\ \midrule
\multirow{4}{*}{Oven}
& PCT & - & - & 4.64  & \textbl{0.73}  & 0.92 & 0.03  & \textbl{0.21}  & 0.73 & 7.46  & 0.49  & 0.72 \\
& ANCSH & 0.69 & 0.46 & 7.11 & 0.51 & 0.74 & 0.31 & 0.02 & 0.02 & 26.50 & 0.08 & 0.17 \\
& CAPT-Plain & \textbl{0.98} & \textbl{0.94}   & \textbl{4.27} & 0.64  & \textbl{0.92}   & \textbl{0.03} & \textbfr{0.22} & \textbl{0.75} & \textbl{6.99}  & \textbl{0.49} & \textbl{0.77} \\
& CAPT (ours) & \textbfr{0.98}& \textbfr{0.94}  & \textbfr{4.22} & \textbfr{0.74} & \textbfr{0.94}  & \textbfr{0.03} & 0.20  & \textbfr{0.76} & \textbfr{5.58} & \textbfr{0.60} & \textbfr{0.87}  \\ \midrule
\multirow{4}{*}{\begin{tabular}[c]{@{}c@{}}Washing\\ Machine\end{tabular}}
& PCT & - & - & 6.77  & 0.44 & 0.85   & 0.04  & 0.22  & 0.77 & 10.40 & 0.36  & 0.63 \\
& ANCSH & 0.83 & 0.56   & 19.10 & 0.21  & 0.35   & 0.30  & 0.04  & 0.05  & 22.60 & 0.11  & 0.23   \\
& CAPT-Plain & \textbl{0.99} & \textbl{0.97}   & \textbfr{5.84} & \textbl{0.52}  & \textbfr{0.91}  & \textbl{0.03}  & \textbl{0.24} & \textbfr{0.95} & \textbfr{8.35} & \textbfr{0.49} & \textbfr{0.75}  \\
& CAPT (ours) & \textbfr{0.99}& \textbfr{0.97}  & \textbl{5.93}  & \textbfr{0.53} & \textbl{0.90}   & \textbfr{0.03} & \textbfr{0.25} & \textbl{0.81}  & \textbl{8.85} & \textbl{0.44}  & \textbl{0.73}\\ \midrule
\multirow{4}{*}{Mean}
& PCT & - & - & 6.31  & 0.48  & 0.85   & 0.04  & 0.20  & 0.71  & 10.23 & 0.22  & 0.37 \\
& ANCSH & 0.73 & 0.52   & 9.60  & 0.43  & 0.70   & 0.26  & 0.03  & 0.14  & 17.66 & 0.04  & 0.09   \\
& CAPT-Plain & \textbl{0.98} & \textbl{0.93}   & \textbl{4.99}  & \textbl{0.59}  & \textbl{0.91}   & \textbl{0.03}  & \textbl{0.26}  & \textbl{0.80}  & \textbl{9.07}  & \textbl{0.26}  & \textbl{0.41}   \\
& CAPT (ours)  & \textbfr{0.98}& \textbfr{0.93}  & \textbfr{4.35} & \textbfr{0.70} & \textbfr{0.95}  & \textbfr{0.03} & \textbfr{0.28} & \textbfr{0.84} & \textbfr{7.53} & \textbfr{0.33} & \textbfr{0.43}  \\ \bottomrule
\end{tabular}
\label{evaluation}
\end{table*}

\subsection{Experimental Results}
\begin{figure}[tb]
	\begin{center}
		\includegraphics[width=6.0cm]{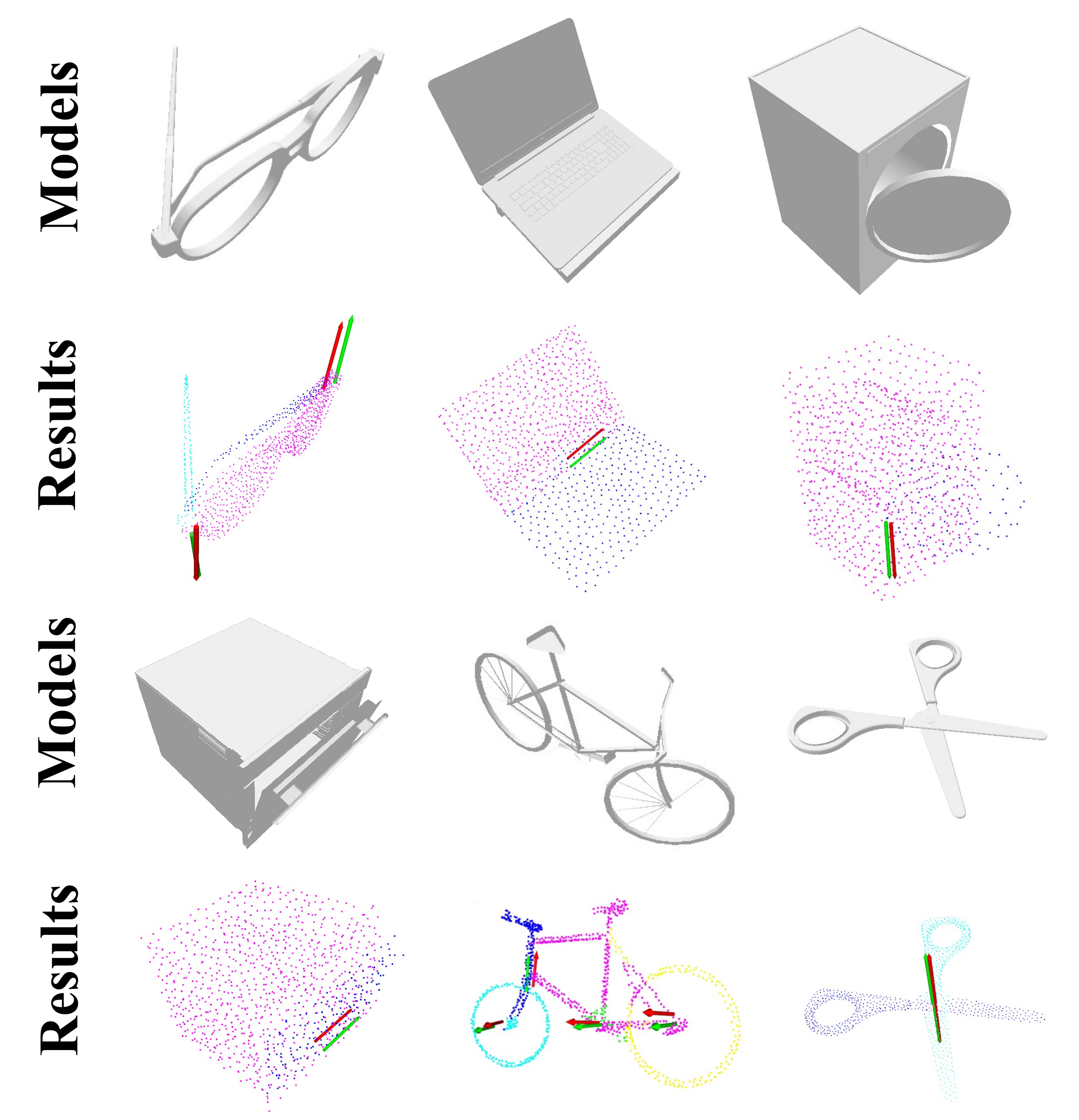}
			\caption[labelInTOC]{Qualitative results for six categories: eyeglasses, laptop, washing machine, oven, bike, and scissors. The models are provided for reference only. The point clouds of washing machine and oven are merged from multiple observations to show the inner structures.}
		\label{vis}
	\end{center}
\end{figure}

Evaluation results are summarized in Table \ref{evaluation}. Qualitative results are visualized in Fig. \ref{vis}. 

For segmentation, our methods achieved a high segmentation accuracy: $>$98\% in all subsets. Moreover, our methods achieved top performance in three of the four estimation categories. Even in the eyeglasses category, our methods obtained results comparable with those of the best-performing alternative method. 

We also determined the influence of motion loss and double voting: for joint parameter estimation, a direct regression method such as PCT might achieve a result slightly better than that of ANCSH, a PointNet- and RANSAC-based method. The comparison of the results between PCT and CAPT-plain shows that a voting-based multi-head decoder is necessary in order to obtain joint parameter estimation of high accuracy and stability. However, CAPT with motion loss and double voting further enhanced precision, especially for joint direction prediction. Double voting also boosted the exactness of the estimation to some extent. 

\begin{table}[tbp]
\centering
\caption[]{Average Time Consumption for Each Input: \textup{CAPT w/o DV means CAPT without using double voting. Laptop represents single-joint objects, while eyeglasses represent multiple-joint objects.}}
\begin{tabular}{@{}llll@{}}
\toprule
Dataset     & ANCSH (s) & CAPT w/o DV (s) & CAPT (s) \\ \midrule
Laptop      & 1.00      & \textbfr{0.035}           & 0.036 \\
Eyeglasses  & 3.33      & \textbfr{0.040}           & 0.041 \\ \bottomrule
\end{tabular}
\label{time}
\end{table}

\begin{figure}[tb]
	\begin{center}
		\includegraphics[width=8.5cm]{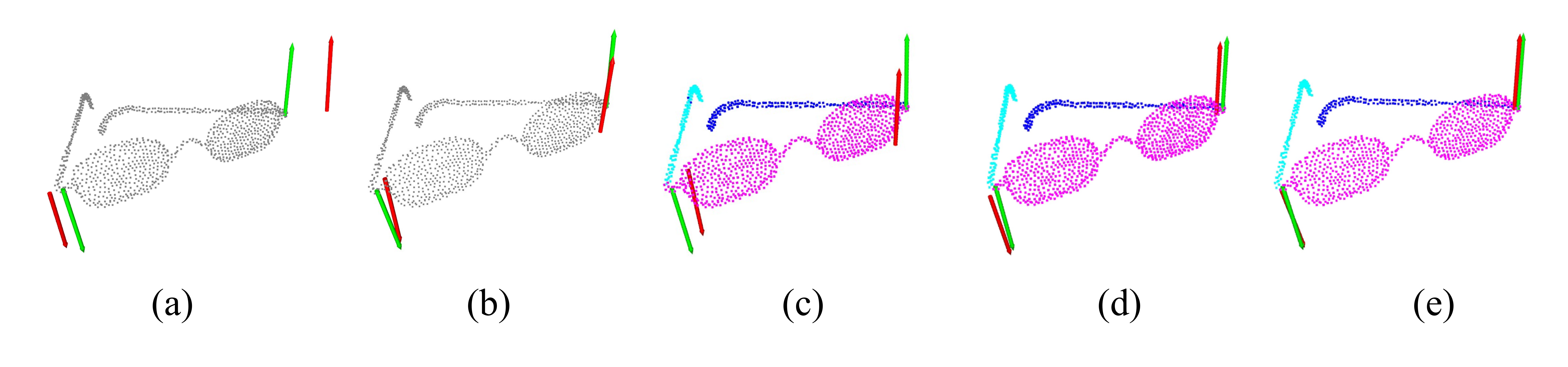}
			\caption[labelInTOC]{Comparisons between (a) ANCSH (b) PCT (c) CAPT-plain (CAPT without double voting and motion loss) (d) CAPT without double voting and (e) CAPT with double voting. Here the object has thin arms, which can make naive PCT unmanageable. On the other hand, our method successfully predicted joint parameter values with relatively high accuracy whether or not double voting was used. Double voting yielded an even better result.}
		\label{compa}
	\end{center}
\end{figure}

The advantage of double voting was strengthened in some hard cases, as shown in Fig. \ref{compa}. It was particularly difficult to conduct successful articulation estimation on objects with such tiny links. These tiny links resulted in (1) few points in the link, thus few structural features, and (2) degenerate surface normals, since each link was reduced to a line with no effective surface. PCT failed in this case, while our method achieved good estimation despite the difficulty. Double voting here filtered out some of the less trustable points, thus obtaining a better result than CAPT-plain. However, we note that the two double voting hyperparameters $\omega_0 and \omega_1$ sometimes needed to be carefully examined before inference in order to avoid loss of accuracy.

\subsection{Time Consumption}
The average time Consumption for each input is shown in Table \ref{time}. Since CAPT's end-to-end structure required no further post-optimization processing, it took a relatively shorter time compared with the multi-stage method. Notably, the inference time consumption of CAPT was relatively insensitive to the joint number of the subset. The increase in inference time between the single-joint laptop subset and the multiple-joint eyeglasses subset was only $13\%$ for CAPT but was up to 2 times more for ANCSH.

\subsection{Direct Simulation-to-reality Result}

\begin{figure}[tb]
	\begin{center}
		\includegraphics[width=8.0cm]{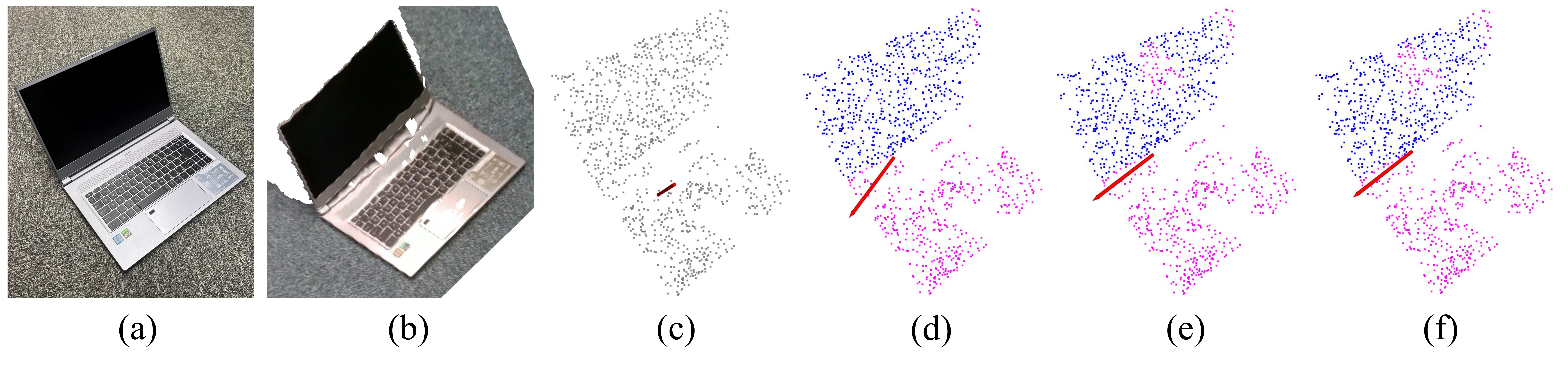}
			\caption[labelInTOC]{Direct sim-to-reality result. (a) Real scene, (b) extracted point cloud, (c) naive PCT, (d) without motion loss, (e) without double voting, and (f) our methods. The results indicate that our category-level articulation estimation from a single point cloud using Transformer (CAPT) methods successfully captured the category features of noisy real-world articulated objects despite being trained with only a synthetic dataset.}
		\label{real}
	\end{center}
\end{figure}

Direct simulation-to-real experiments indicated that the CAPT model exhibits promising performance in real-world articulation estimation, even in the absence of fine-tuning on real-world datasets. This finding is demonstrated by the direct simulation-to-reality results presented in Fig. \ref{real}. Prior to inputting the real-world point cloud data into the model, we conducted a series of pre-processing steps, including the use of RANSAC to fit and remove the ground plane; statistical outlier removal to denoise the point cloud; and point cloud normalization \cite{guoFastRobustBinpicking2020}.

\section{CONCLUSION}
In this paper, we proposed a set of articulation estimation methods: articulation point transformer, with a motion loss approach to recover dynamic features from a single static point cloud, and coarse-to-fine double voting to achieve high accuracy in category-level articulation estimation. We demonstrated better performance for these methods than for existing alternative methods or for baseline performance on a multi-category articulated object data set. The end-to-end structure of our study distinguishes it from previous work in which post-optimization or multi-stage networks were used. Future work will focus on cross-category articulation estimation with less or no prior kinematic constraint knowledge required. We are also working on the deployment of this articulation estimation system in real-scene robot manipulation, as a promising application.




\section*{ACKNOWLEDGMENT}

This work was partly supported by the social cooperation program "Technology for IoT sensing and analysis," sponsored by UTokyo and Air Water.



\IEEEtriggeratref{20}

\bibliography{myref}
\bibliographystyle{IEEEtran}

\end{document}